\documentclass[fleqn,10pt]{SelfArx} 

\usepackage[english]{babel} 

\usepackage{lipsum} 

\definecolor{color1}{RGB}{0,0,90} 
\definecolor{color2}{RGB}{0,20,20} 

\usepackage{hyperref} 

\hypersetup{
	hidelinks,
	colorlinks,
	breaklinks=true,
	urlcolor=color2,
	citecolor=color1,
	linkcolor=color1,
	bookmarksopen=false,
	pdftitle={Title},
	pdfauthor={Author},
}

\Archive{Pre-Print} 

\PaperTitle{Real-World Performance of Autonomously Reporting Normal Chest Radiographs in NHS Trusts Using a Deep-Learning Algorithm on the GP Pathway} 

\Authors{J. Smith\textsuperscript{1}*, T. Naunton Morgan\textsuperscript{1}, P. Williams\textsuperscript{1,2}, Q. Malik\textsuperscript{1,3}, S. Rasalingham\textsuperscript{1}} 
\affiliation{\textsuperscript{1}\textit{behold.ai Technologies Limited, 123 Buckingham Palace Road, London SW1W 9SH}} 
\affiliation{\textsuperscript{2}\textit{Department of Radiology, University Hospitals Coventry and Warwickshire, Coventry, UK}}
\affiliation{\textsuperscript{3}\textit{Department of Radiology, Mid and South Essex NHS Trust, Essex, UK}} 
\affiliation{*\textbf{Corresponding author}: J. Smith (jordans@behold.ai)} 

\Abstract{AIM: To analyse the performance of a deep-learning (DL) algorithm currently deployed as diagnostic decision support software in two NHS Trusts used to identify normal chest x-rays in active clinical pathways.\\ \\
MATERIALS AND METHODS: A DL algorithm has been deployed in Somerset NHS Foundation Trust (SFT) since December 2022, and at Calderdale \& Huddersfield NHS Foundation Trust (CHFT) since March 2023. The DL algorithm was developed and trained prior to deployment, and is used to assign abnormality scores to each GP-requested chest x-ray (CXR). The algorithm then classifies a subset of examinations with the lowest abnormality scores as High Confidence Normal (HCN), and displays this result to the Trust. This two-site study includes 4,654 CXR continuous examinations processed by the algorithm over a six-week period.\\ \\
RESULTS: When classifying 20.0\% of assessed examinations as HCN, the model classified exams with a negative predictive value (NPV) of 0.96. There were 0.77\% (36) of examinations classified incorrectly as HCN, with none of the abnormalities considered clinically significant by auditing radiologists. The DL software maintained fast levels of service to clinicians, with results returned to Trusts in a mean time of 7.1 seconds.  \\ \\
CONCLUSION:The DL algorithm performs with a low rate of error and is highly effective as an automated diagnostic decision support tool, used to autonomously report a subset of CXRs as normal with high confidence. Removing 20\% of all CXRs reduces workload for reporters and allows radiology departments to focus resources elsewhere.}

\begin{document}

\twocolumn[{%
		\thispagestyle{empty}%
		\vskip-36pt%
		{\raggedleft\small\sffamily\bfseries Pre-Print\\}%
		\vskip20pt%
		{\raggedright\color{color1}\sffamily\bfseries\fontsize{20}{25}\selectfont Real-World Performance of Autonomously Reporting Normal Chest Radiographs in NHS Trusts Using a Deep-Learning Algorithm on the GP Pathway\par}%
		\vskip10pt%
		{\raggedright\color{color1}\sffamily\fontsize{12}{16}\selectfont J. Smith\textsuperscript{1}*, T. Naunton Morgan\textsuperscript{1}, P. Williams\textsuperscript{1,2}, Q. Malik\textsuperscript{1,3}, S. Rasalingham\textsuperscript{1} \par}%
		\vskip18pt%
		\fcolorbox{color1}{white}{%
			\parbox{\textwidth-2\fboxsep-2\fboxrule}{\centering%
				\colorbox{color2!10}{%
					\parbox{\textwidth-4\fboxsep-2\fboxrule}{%
							\sffamily\textbf{\abstractname}\\AIM: To analyse the performance of a deep-learning (DL) algorithm currently deployed as diagnostic decision support software in two NHS Trusts used to identify normal chest x-rays in active clinical pathways.\\ \\
MATERIALS AND METHODS: A DL algorithm has been deployed in Somerset NHS Foundation Trust (SFT) since December 2022, and at Calderdale \& Huddersfield NHS Foundation Trust (CHFT) since March 2023. The DL algorithm was developed and trained prior to deployment, and is used to assign abnormality scores to each GP-requested chest x-ray (CXR). The algorithm then classifies a subset of examinations with the lowest abnormality scores as High Confidence Normal (HCN), and displays this result to the Trust. This two-site study includes 4,654 CXR continuous examinations processed by the algorithm over a six-week period.\\ \\
RESULTS: When classifying 20.0\% of assessed examinations as HCN, the model classified exams with a negative predictive value (NPV) of 0.96. There were 0.77\% (36) of examinations classified incorrectly as HCN, with none of the abnormalities considered clinically significant by auditing radiologists. The DL software maintained fast levels of service to clinicians, with results returned to Trusts in a mean time of 7.1 seconds.  \\ \\
CONCLUSION: The DL algorithm performs with a low rate of error and is highly effective as an automated diagnostic decision support tool, used to autonomously report a subset of CXRs as normal with high confidence. Removing 20\% of all CXRs reduces workload for reporters and allows radiology departments to focus resources elsewhere.
					}%
				}%
				\vskip4pt%
				\begingroup%
					\raggedright\sffamily\small%
					\footnotesize\textsuperscript{1}\textit{behold.ai Technologies Limited, 123 Buckingham Palace Road, London SW1W 9SH}\par%
                    \textsuperscript{2}\textit{Department of Radiology, University Hospitals Coventry and Warwickshire, Coventry, UK}\par%
                    \textsuperscript{3}\textit{Department of Radiology, Mid and South Essex NHS Trust, Essex, UK}\par%
 
                    *\textbf{Corresponding author}: J. Smith (jordans@behold.ai)\par%
				\endgroup
			}%
		}%
		\vskip25pt%
	}]%

\thispagestyle{empty} 


\section*{Introduction} 

Between April 2021 and April 2022, 7.9 million chest x-rays (CXRs) were performed in the NHS \cite{NHSEng:2020}, up from 6.7 million between April 2020 and April 2021. CXRs are the most common diagnostic imaging test \cite{table_4} and are the most common test requested by GPs, having increased in frequency by 67\% compared to 2020/21 \cite{did_2022}.

Radiology departments are under high levels of pressure to deal with a growing patient backlog. More patients than ever are waiting for treatment, with 7.3 million patients on NHS waiting lists for care as of March 2023. These figures have grown since the pre-pandemic era, in which 4.4 million patients were on waiting lists as of February 2020 \cite{bma_backlog}. As of March 2023, the median waiting time for treatment was 14.1 weeks. This is double the pre-pandemic median wait time of 6.9 weeks as of March 2019 \cite{nhs_waiting_times}. Cancer targets continue to be missed; in March 2023, 71.6\% of patients received their first treatment within two months of attending a screening service, falling below the operational standard of 90\% \cite{bma_backlog}. 

These pressures are compounded by a national shortfall in radiologists \cite{rcr_census}. The NHS radiologist workforce is short-staffed by 33\% and is forecast to be short-staffed by 39\% by 2026 \cite{rcr_cuts}, with 58\% of radiology leaders claiming they do not have enough diagnostic and interventional radiologists to keep patients safe \cite{rcr_backlogs}. 

To deal with these growing pressures, the Royal College of Radiologists has called for ‘significant investment in modern technology to support early detection and treatment’ \cite{rcr_backlogs}. The use of DL algorithms to augment radiology pathways has drawn recent attention. Several studies have been published in this domain, including those which evaluate the effect of deep learning on radiologist accuracy \cite{seah_2021}, or to compare deep learning approaches for CXR classification \cite{baltruschat_2019}. However, the majority of published studies measured the performance of these algorithms using artificial datasets, which differ from real-world conditions in terms of disease prevalence, population diversity and image quality. These studies often demonstrate the isolated performance of models, rather than assessing a model’s real-world value as part of active clinical pathways within the NHS.

A DL algorithm has been deployed commercially in two NHS Trusts and is used for autonomous diagnostic decision support. The algorithm is a UKCA Class IIa device (red dot ® v2.2, Behold.ai), meaning it conforms to the requirements stated by UK Medical Devices Regulation 2002 \cite{gov_legislation}, meets legislation relating to safety \cite{ukca} and is able to issue an authorised clinical report. It has been deployed in Somerset Foundation Trust (SFT) since December 2022, and at Calderdale \& Huddersfield NHS Trust (CHFT) since March 2023. The clinical use of this DL algorithm is to issue a fully authorised clinical diagnosis report for examinations that it identifies as being normal with a high degree of confidence. 

To maintain its efficacy, the DL algorithm must demonstrate the capacity to flag a significant proportion of exams as HCN with a low rate of error in real-world clinical settings. The algorithm is used to provide a rule-out test for HCN on clinical CXRs, so as not to require further human interpretation or intervention in the clinical pathway. The objective of this study is to evaluate the performance and efficacy of this DL algorithm.


\section{Methods}

\subsection{Study Design}
This study serves to analyse and validate the performance of a DL algorithm to detect normality in adult CXRs and to issue a clinical report to that effect. Each examination was assessed by the algorithm and an abnormality score produced. Those with the lowest scores were classified as normal with a high degree of confidence. These examinations are labelled as ‘high confidence normal’ (HCN), and an automated flag is returned to the Trust to identify examinations as such. Radiology reports are autonomously generated for these exams, thus removing them from radiologists’ workload. 

\subsection{Data Sources}
A total of 4,076 patients accounting for a total of 4,654 radiographs from two NHS Trusts were used for evaluation, with 2,013 radiographs from SFT and 2,641 radiographs from CHFT. This study uses all processed CXR data collected from a continuous six-week period across both sites, ranging from 1st April 2023 to 13th May 2023. Differences in population age were determined to be statistically insignificant using Student’s t-test (p $<$ 0.05). This study used fully anonymised radiographs and as such there was no requirement to obtain informed consent from individuals.

\begin{table}[hbt]
    \small
	\caption{Study Population Distribution \\ Figures for patient sex are (female/male/not specified)}
	\centering
	\begin{tabular}{ccccc}
		\toprule
		     -  & Total CXRs & Patients & Avg. Age & Sex (F/M/NS) \\
		\midrule
		SFT    & 2013   & 2001 & 63.4 & 1071/930/0  \\
		CHFT   & 2641 & 2075 & 60.9 & 1134/940/1 \\
            Total & 4654 & 4076 & 62.1 & 2205/1870/1 \\
		\bottomrule
	\end{tabular}
	\label{tab:demographics}
\end{table}

This study uses the real-world continuous dataset collected across two NHS trusts over a six-week period. The dataset has not been artificially adjusted to optimise the balance of patient subgroups included in the analysis, and thus the performance is reflective of the DL algorithm’s deployment in active NHS GP referral pathways. 

Both posteroanterior (PA) and anteroposterior (AP) chest radiographs were included. In addition to the pixel data, the following non-identifying information was collected: patient age, patient sex, AP or PA, radiography unit, and technique (computed radiography [CR] and digital radiographs [DR]).

These images were produced using at least 7 different devices, made by 5 different manufacturers across the two sites (Table \ref{tab:manufacturers}). The techniques present in the dataset were computed radiography (CR), 60.9\% and digital radiographs (DR), 39.1\%. 

\begin{table}[hbt]
    \small
	\caption{CXR Manufacturers}
	\centering
	\begin{tabular}{cccccc}
		\toprule
		     - & Canon & Samsung & Philips & Fujifilm & Kodak \\
		\midrule
		SFT    & $833$   & $970$ & $178$ & $32$  & $0$ \\
		CHFT   & $885$   & $6$   & $0$   & $484$ & $1266$ \\
            Total  & $1718$  & $976$ & $178$ & $516$ & $1266$ \\
		\bottomrule
	\end{tabular}
	\label{tab:manufacturers}
\end{table}

Criteria for inclusion in this study were frontal CXRs performed on adults $\geq$ 18 years of age, of technique CR or DR, and referred by general practitioner (GP). Criteria for exclusion were non-CXR examinations including lateral CXR and chest-abdomen radiographs. In order to comply with the requirements of EU GDPR and The Data Protection Act 2018, personal patient information in the DICOM header meta-data were anonymised to de-identify the patients.

\subsection{Definition of Normality}
In the scope of this study, a normal adult chest radiograph is defined by the following criteria: a frontal image performed in inspiration showing a well-penetrated radiograph. Vertebrae are visible behind the heart. Left hemidiaphragm is visible to the edge of the spine. The lungs are appropriately visualised, and the vascular markings are not prominent. The entire chest including lung apices and the costophrenic angles should be included in the field of view. Absence of abnormality in the lungs area, the mediastinum, pleural space, bones and upper abdomen, except for mild scoliosis in patients over 35 years of age. Absence of medical devices, except for electrocardiogram leads and clips. In addition to this, any radiographs with insufficient quality to determine the above criteria were designated as ‘sub-optimal’ by radiologists and considered ‘abnormal’ examinations for the purpose of this study. This definition of normal is in line with previously published analyses of this DL algorithm's performance \cite{dyer_2021}.

\subsection{Source of Audit Labels}
\begin{figure}
    \centering
    \includegraphics[width=0.41\textwidth]{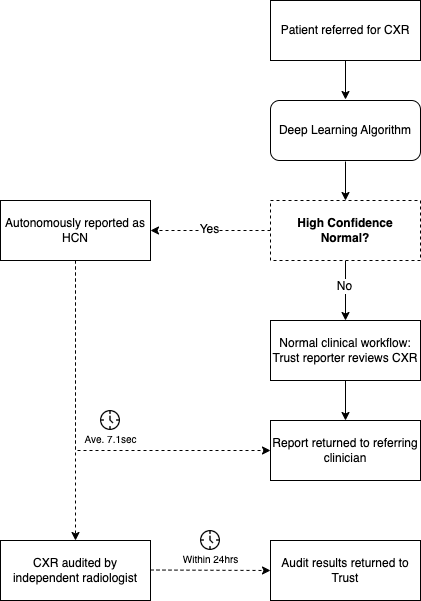}
    \caption{Workflow of autonomous diagnostic decision support by rule-out HCN, as deployed in NHS Trusts}
    \label{fig:hcn_workflow}
\end{figure}

This study uses performance data from the real-world deployment of DL software for diagnostic decision support, and as such reflects the single-reader approach used in the active clinical pathway. Following the definition of normality, each image classed as HCN was reviewed and labelled by one of three independent FRCR trained radiologists. Two of these radiologists have a minimum of 10 years NHS clinical experience at consultant level, and the other has 6 years clinical experience across both the NHS and Indian healthcare. Of the examinations classed as HCN and subsequently reviewed during this time period, 84.4\% were audited by Radiologist 1, 8.0\% were audited by Radiologist 2 and 7.6\% were audited by Radiologist 3. The results of this audit process are communicated to the relevant Trust within 24 hours of the radiograph's submission. This workflow is detailed in Fig \ref{fig:hcn_workflow}.

\subsection{Deep Learning Algorithm}
The DL algorithm used in this study is a commercial medical device (red dot® v2.2, Behold.ai). It is an ensemble of individually trained convolutional neural networks (CNNs), with the final predictions being an average of each algorithm’s score. The ensemble contained a mix of two different types of model architecture, DenseNet-121 \cite{Huang_2017_CVPR} and EfficientNet-B4 \cite{pmlr-v97-tan19a}. The individual models were trained on datasets of approximately 150,000 chest radiographs collected from the National Institutes of Health (NIH) and the NHS. Although data coming from the institutions in this study was used in model training, there is no overlap with the examinations used in this study and those used for training.

Following its collection, model training data was labelled by radiologists. A large proportion of the training data was labelled using a ‘ground-truthing’ process in which two FRCR trained radiologists independently label the images, with a third radiologist arbitrating any cases of disagreement between them. Many of the radiographs used for training were annotated by radiologists with either bounding boxes or freehand annotations to denote the region of interest for the labelled pathology. In total 45 different abnormal radiograph classes were used in the training process, with the absence of any labels indicating a normal case. 

The model was developed using Python 3.5 and the PyTorch library, and processes pixel-level data only in its prediction of abnormality. The software also has the functionality to prioritise lung cancer findings present on CXRs, which is outside the scope of this study.

\subsection{HCN Output}
After processing a radiograph, the DL algorithm outputs a continuous abnormal score between 0 and 1 to classify the image as either normal (0.0) or abnormal (1.0). This score relates to the probability that the image is abnormal according to the model and is also referred to as the ‘confidence score’. The binary output is obtained by comparing the score to a pre-determined threshold, whereby examinations with an abnormal score below the threshold are classified as HCN.

To optimise the model’s ability to classify examinations as normal, site-specific HCN thresholds were calculated for each Trust. This approach to site-specific calibration as a means to boost performance was discussed in Dyer, et al (2021) \cite{dyer_2021}. The site-specific thresholds were determined using separate datasets from SFT and CHFT, referred to as ‘calibration datasets’. These sets of exams were collected before deployment at each Trust to act as a representative sample of radiographs from which to calculate the threshold. The calibration dataset from SFT contained 1,004 radiographs, collected between 15th July 2021 and 10th August 2021. The calibration dataset from CHFT contained 984 radiographs collected between 1st October 2022 and 9th November 2022. Radiographs contained in these calibration sets are only used to calculate respective HCN thresholds, and have not been included in model training or the contents of this study. 

Although the proportion of exams classified as HCN differed in live deployment between the two sites, the site-specific thresholds have been retrospectively adjusted to rule out 20\% of examinations as HCN for the purposes of this study. 

\subsection{Statistical Analysis}
To evaluate and validate the algorithm performance, negative predictive value (NPV) and discrepancy rate were measured using the HCN output of the model. NPV corresponds to the proportion of HCN cases detected by the model that were normal CXRs according to auditing radiologists. Discrepancy rate corresponds to the proportion of all processed exams that were incorrectly classified as HCN according to auditing radiologists, with uncertainty quantified using 95\% confidence intervals and p-values calculated using Student's t-test.


\section{Results}

\subsection{Model Performance}

The study assessed the DL algorithm’s performance on classifying HCN examinations in an active clinical pathway. Of the 4654 radiographs processed across both sites, the algorithm classified examinations as HCN with a negative predictive value (NPV) of 0.96 and a discrepancy rate of 0.77\% with 95\% confidence interval +/- 0.0025 (p $<$ 0.05). The radiographs classified as HCN included 36 abnormal examinations, which were considered to have been incorrectly classified as normal. Of these 36 discrepancies among the 930 radiographs classified as HCN, none were considered to be ‘clinically significant’ by auditing radiologists, where the term ‘clinically significant’ denotes a finding that could potentially be indicative of a serious illness (Fig. \ref{fig:significant_findings}).

\begin{figure}
    \raggedright
    \begin{tabular}{l}
        \textbf{List of query significant findings on chest radiographs} \\
        \midrule
         Collapse \\
         Hilar enlargement (lymph, cancer) \\
         Mass $>$ 3cm \\
         Nodule $<$ 3cm \\
         Pleural effusion \\
         Pneumothorax \\
         Pulmonary oedema \\
         Subcutaneous emphysema \\

    \end{tabular}
    \caption{Query Significant Findings}
    \label{fig:significant_findings}
\end{figure}

At SFT, the algorithm classified 403 out of 2,013 examinations as HCN with an NPV of 0.96 and a discrepancy rate of 0.79\% with 95\% confidence interval +\- 0.0039 (p $<$ 0.05). Of the 403 HCN exams, 16 were discovered to contain abnormalities. 

At CHFT, the DL algorithm classified 527 out of 2,641 examinations as HCN with an NPV of 0.96 and a discrepancy rate of 0.76\% with 95\% confidence interval +/- 0.0033 (p $<$ 0.05). Of these 527 exams, 20 were incorrectly classified as normal.

\subsection{HCN Discrepancies}

\begin{figure}
    \centering
    \includegraphics[width=0.5\textwidth]{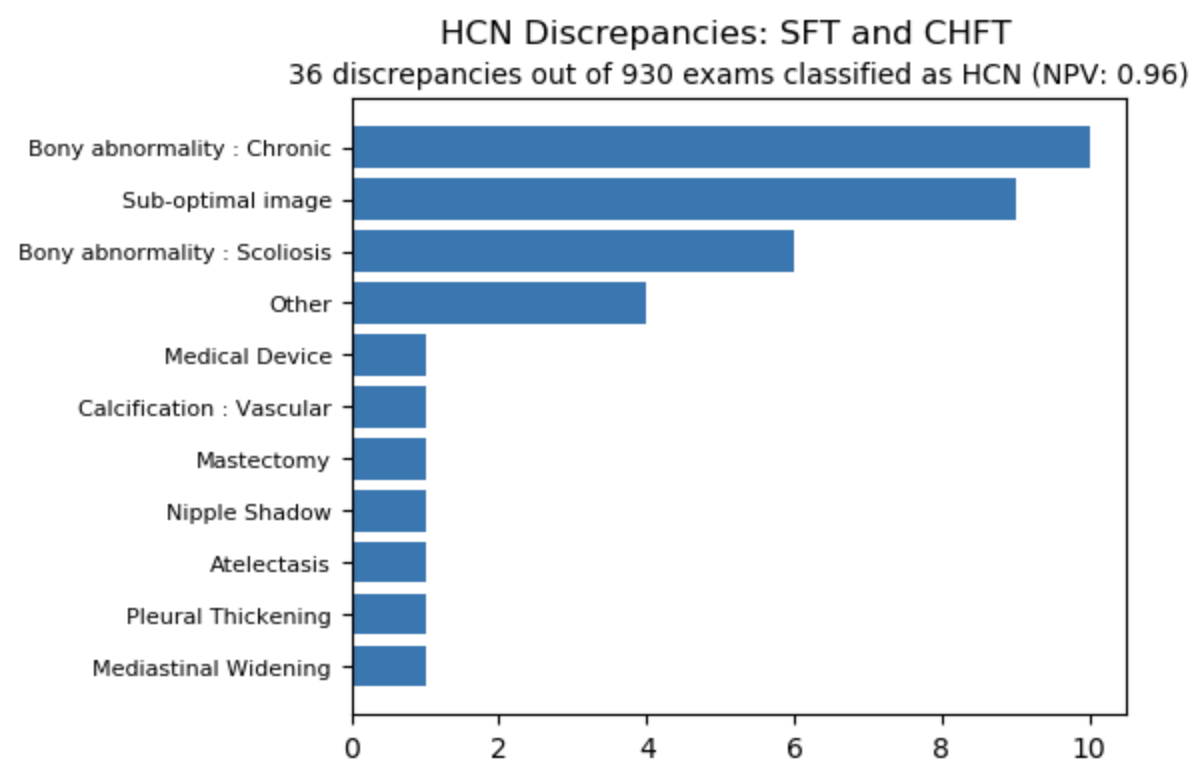}
    \caption{HCN Discrepancies}
    \label{fig:hcn_results}
\end{figure}

\begin{figure}
    \centering
    \includegraphics[width=0.5\textwidth]{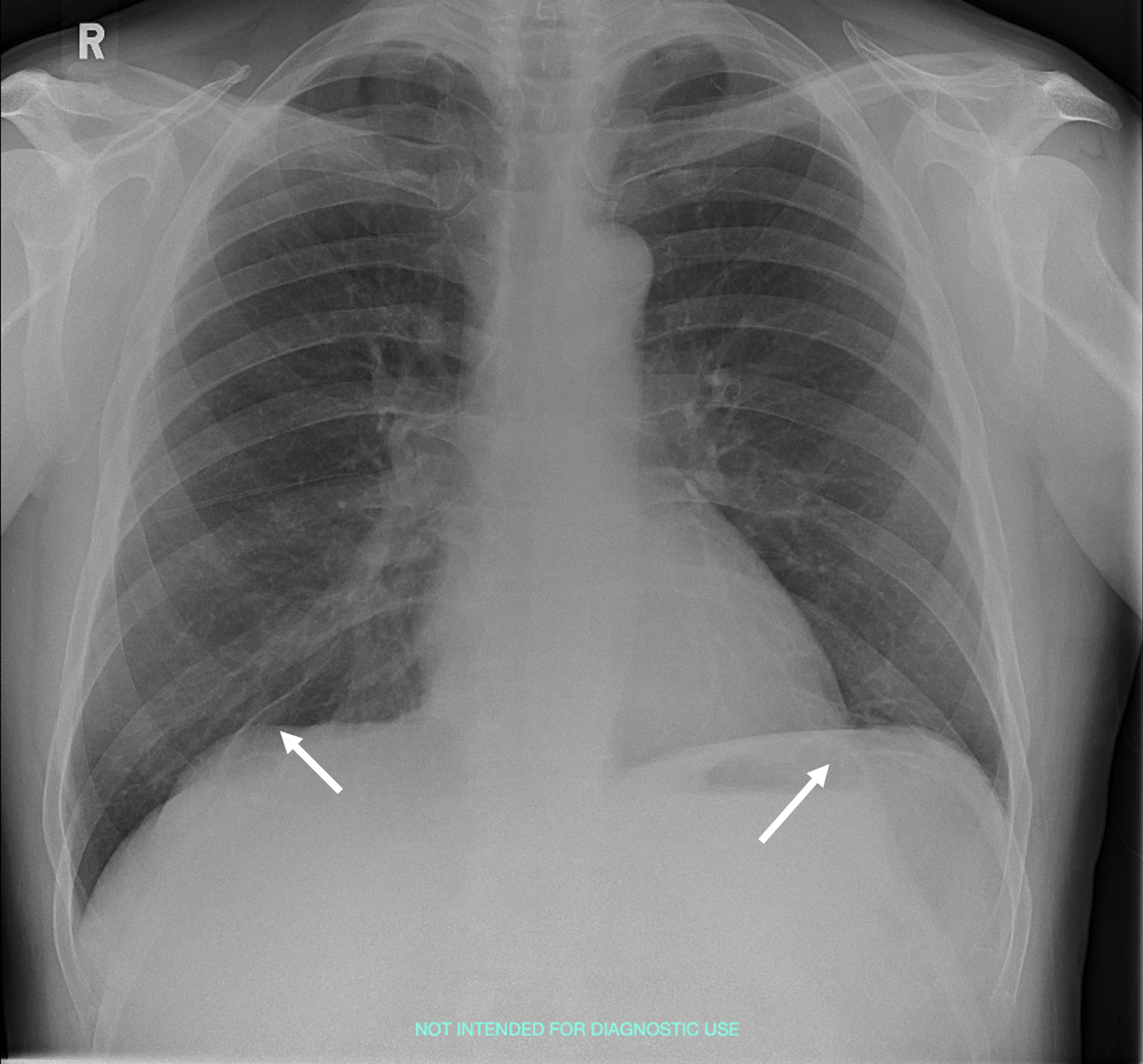}
    \caption{Atelectasis incorrectly classified as HCN by the DL algorithm}
    \label{fig:atelectasis}
\end{figure}

The distribution of abnormalities across both sites can be found in fig. \ref{fig:hcn_results}, and an example of a misclassified CXR can be found in fig. \ref{fig:atelectasis}. Overall, the most common category of abnormality found in misclassified exams was chronic bony abnormalities, with 6 originating from CHFT and 4 from SFT. These were highlighted by auditing radiologists as containing congenital rib abnormalities, normal variants such as cervical ribs, old rib fractures or sclerotic foci.

The joint-second most common category of abnormality was sub-optimal images, of which there were 9 misclassified as normal. Though a sub-optimal image does not indicate the presence of an abnormality, sub-optimal images are radiographs with insufficient quality to determine normality to a high degree of confidence and as such are classed as abnormal for the purpose of this study. Of the 9 misclassified sub-optimal images, 7 originated from SFT and 2 originated from CHFT. Of these 9 sub-optimal images, 4 contained missing costophrenic angles, 3 contained missing lung apices and 2 were rotated.

Radiographs labelled as ‘other’ were identified as containing abnormalities which did not fit the existing abnormality classes. Of the 4 misclassified radiographs given this label, 3 originated from CHFT and contained azygos lobe, and 1 originated from SFT containing a deviated trachea. 

\subsection{Service Levels}

Service levels throughout the study period remained high. Upon either site submitting a radiograph, results of the DL algorithm were returned in a mean time of 7.1 seconds (range 5-17 seconds) 

Radiographs classed as HCN by the DL algorithm were audited by independent radiologists and results were returned to the hospital. Of these exams, 99.3\% were audited within 24 hours of the radiograph’s submission. The average time taken from exam submission to audit was 3 hours and 50 minutes.

\section{Discussion}

This study highlights an effective framework for combining DL decision support software with human expertise. Though none of the discrepancies present during the timeframe of study were considered to be clinically significant upon review, the presence of independent radiologists auditing the HCN exams acted as an important means of identifying abnormalities misclassified by the algorithm. The prompt return of audit results to NHS Trusts (99.3\% within 24 hours of exam submission) provided assurance to clinicians and served to build levels of trust in the DL algorithm’s performance.
 
 The success of the framework used in this study lends support to a growing consensus in scientific literature, that artificial intelligence should be used to complement radiologists rather than replacing them \cite{phillips_2022}. The results of this study support the proposition that radiologist expertise should be combined with artificial intelligence tooling to have the best effect on clinical pathways \cite{leibig_2022}.

\subsection{Model Robustness}

In the field of medical computer vision, many models have been shown to learn hospital-specific configurations and artefacts \cite{antun_2020}, or to amplify biases present in a given institution \cite{robinson_2020}, \cite{gichoya_2022}. Recent studies have highlighted the need for multi-site evaluations of medical DL algorithms \cite{wu_2021}, and regulatory bodies have emphasised the requirement for analysing performance across patient demographics and subgroups \cite{gov_2021}. Disparities in cross-site performance can indicate that a DL algorithm may not be equitable or effective when deployed on diverse patient populations.

Performance of the DL algorithm was similar across both sites. At SFT the algorithm classified examinations with an NPV of 0.96 and a discrepancy rate of 0.79\%, and at CHFT the algorithm removed exams with an NPV of 0.96 and a discrepancy rate of 0.76\%. This common level of accuracy indicates the robustness of the DL algorithm when deployed in environments that differ to the training and in-house testing environments. It also indicates its equitability and effectiveness when deployed across different geographies, equipment types and patient demographics. This real-world performance confirms the high levels of robustness indicated by Dyer, et al (2022) \cite{dyer2022robustness} in which the equitability of this DL algorithm was tested on an artificial validation set across different patient and environmental subgroups. 

\subsection{Further study}
For further insight into the DL model’s performance, researchers may wish to subject the full dataset used in this study to a ground-truthing process, in which each CXR is reviewed by two independent radiologists and disagreements are arbitrated by a third. This would provide insight into disagreement rates between radiologists and would highlight the subjectivity of reporting in comparison to a DL algorithm. It would also indicate which of the HCN discrepancies contained abnormalities subtle enough to be missed by at least one radiologist. 

Labelling all processed radiographs, rather than the subset of exams classified as HCN, would allow the calculation of additional metrics such as precision and recall. Such insights would be useful when comparing the performance of the DL algorithm to that of clinical radiologists, however this study reflects the real-world performance of DL software and as such additional labelling is outside the present scope.

\begin{figure}
    \centering
    \includegraphics[width=0.5\textwidth]{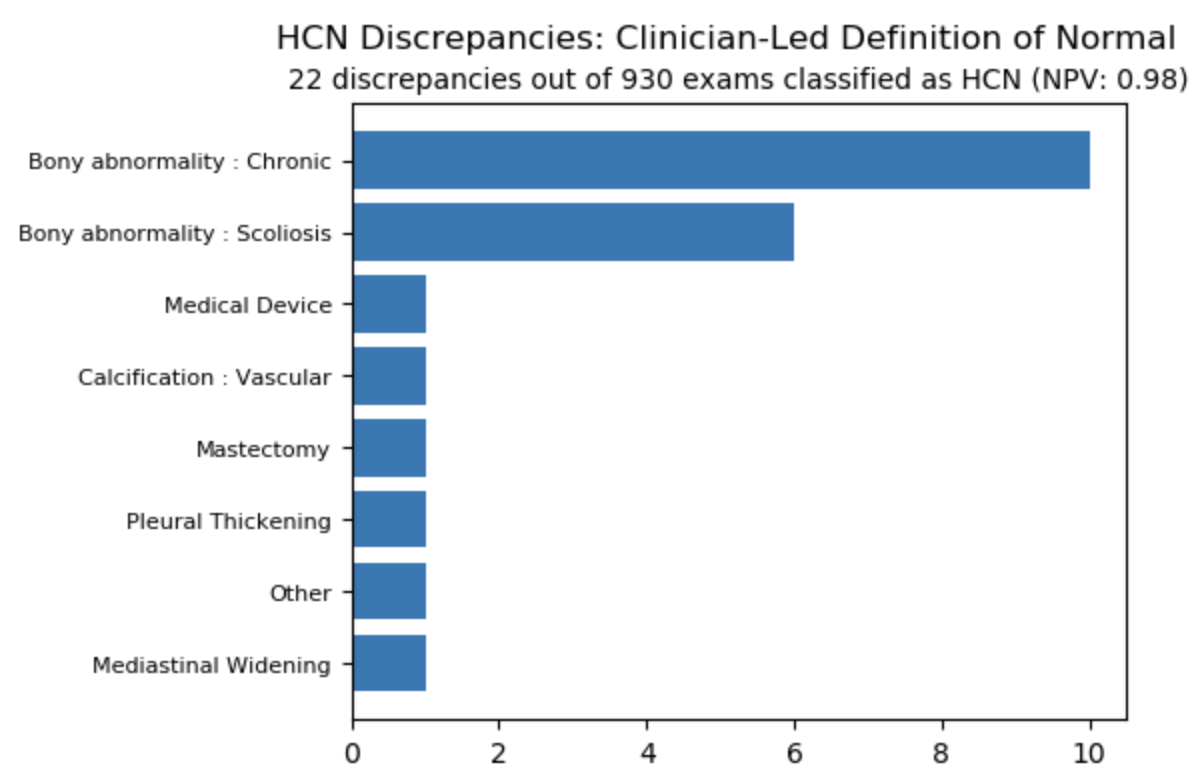}
    \caption{HCN discrepancies with clinician-led definition of normal}
    \label{fig:clinician_led_hcn_results}
\end{figure}

Feedback from study sites indicates that many of the discrepancies flagged by independent radiologists are  unlikely to be of clinical relevance to reviewing clinicians. These clinically insignificant abnormalities include mild congenital anomalies, normal anatomical variants, marginally sub-optimal images, small chronic scars, old rib fractures and certain other benign findings. They have indicated that many of these findings, in practice, would not have been flagged as abnormal by reporting clinicians. By defining ‘normal’ in a manner more reflective of real-world  diagnoses, the number of discrepancies within this study decreases from 36 to 22. In this manner, algorithm performance improves from an NPV of 0.96 to 0.98, and from a discrepancy rate of 0.77\% to 0.47\%. Per-site performance remains highly similar, with an NPV of 0.98 at both SFT and CHFT and discrepancy rates of 0.45\% and 0.49\% respectively. The adjusted distribution of HCN discrepancies can be found in fig \ref{fig:clinician_led_hcn_results}.

This approach further reduces workload for radiologists, who will be tasked with reviewing fewer exams containing minor discrepancies that require no further action. This methodology will be considered in future live deployments.

\subsection{Summary}
This study has shown that a DL algorithm can be successfully used for autonomous diagnostic decision support in active GP referral pathways, removing a significant proportion of normal examinations from the radiology workflow with a low rate of error.

Performance was similar at both NHS Trusts, demonstrating that the DL algorithm generalises across different populations, geographies and equipment types. This study also highlights the success of its technical integration into the NHS, with fast levels of service maintained throughout the study period.

\phantomsection
\section*{Acknowledgments} 
The authors thank both Somerset NHS Foundation Trust and Calderdale \& Huddersfield NHS Foundation Trust for their support in deploying this DL diagnosis decision support software, and for their ongoing feedback on its implementation. The authors thank Dr Adrian Hood for his thoughtful comments and insightful feedback on the contents of this paper. 

The authors also thank Apollo Radiology International for their support in auditing CXR data.  

\phantomsection
\section*{Declaration of Interests}
J. Smith, T. Naunton Morgan, P. Williams, Q. Malik and S. Rasalingham are employed by behold.ai and/or have stock or stock options in behold.ai. No other relevant relationships were disclosed.

\phantomsection
\bibliographystyle{unsrt}
\bibliography{HCN_Paper.bib}


\end{document}